%% file: MAIN.tex
\documentclass[runningheads]{llncs}

\usepackage{eccv}

\usepackage{eccvabbrv}

\usepackage{graphicx}
\usepackage{booktabs}
\usepackage{amsmath}
\usepackage{amssymb}
\usepackage{bm}

\let\svthefootnote\thefootnote
\newcommand\freefootnote[1]{%
  \let\thefootnote\relax%
  \footnotetext{#1}%
  \let\thefootnote\svthefootnote%
}

\usepackage[accsupp]{axessibility}  

\usepackage{hyperref}

\usepackage{orcidlink}

\newcommand{\MethodName}{TalkinNeRF\xspace}

\newcommand{\mytilde}{\raise.17ex\hbox{$\scriptstyle\mathtt{\sim}$}}

\begin{document}

\title{\MethodName: Animatable Neural Fields for Full-Body Talking Humans}

\titlerunning{\MethodName}


\author{Aggelina Chatziagapi\inst{1, *} \and Bindita Chaudhuri\inst{3, *} \and Amit Kumar\inst{2} \and \\Rakesh Ranjan\inst{2} \and Dimitris Samaras\inst{1} \and Nikolaos Sarafianos\inst{2}
}

\authorrunning{A.~Chatziagapi et al.}

\institute{Stony Brook University \email{\{aggelina,samaras\}@cs.stonybrook.edu}
\and Meta Reality Labs \email{\{akumar14,rakeshr,nsarafianos\}@meta.com}
\and Flawless AI \email{bindita.chaudhuri@flawlessai.com}}

\maketitle

\begin{center}
    \captionsetup{type=figure}
    \includegraphics[width=\textwidth]{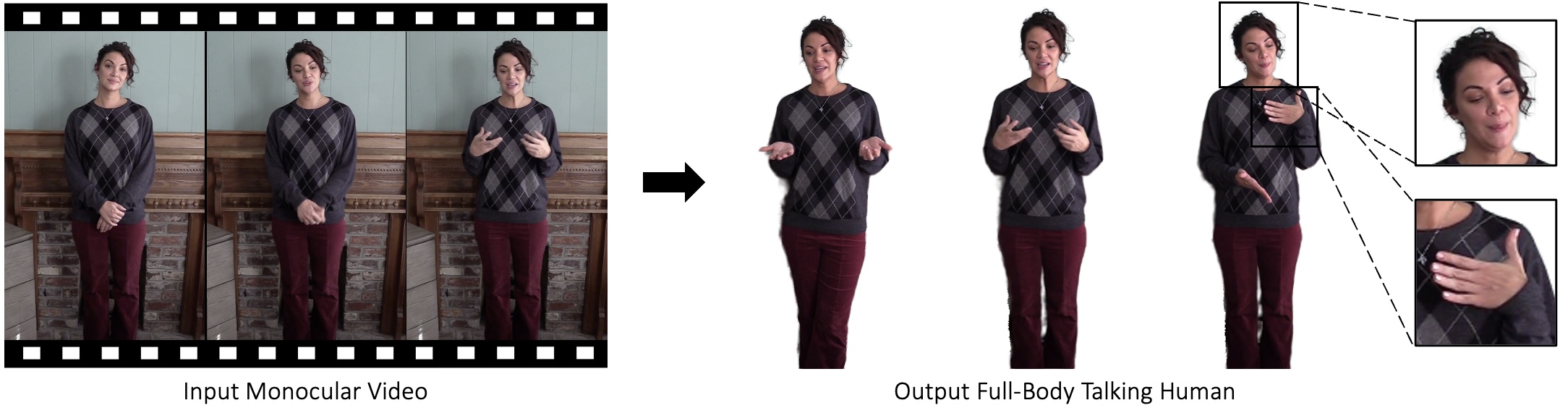}
    \vspace{-5mm}
    \caption{Given monocular videos, \MethodName learns a unified NeRF-based network that represents the \emph{holistic} 4D human motion, including body pose, hand articulation, and facial expressions. It synthesizes high-quality animations of \emph{full-body talking humans}.}
\label{fig:teaser}
  \vspace{-0.3cm}
\end{center}

\begin{abstract}
We introduce a novel framework that learns a dynamic neural radiance field (NeRF) for full-body talking humans from monocular videos. 
Prior work represents only the body pose or the face. However, humans 
communicate with their full body, combining body pose, hand gestures, as well as facial expressions.
In this work, we propose \MethodName, a unified NeRF-based network that represents the holistic 4D human motion.
Given a monocular video of a subject, we learn corresponding modules for the body, face, and hands, that are combined together to generate the final result. To capture complex finger articulation, we learn an additional deformation field for the hands. Our multi-identity representation enables simultaneous training for multiple subjects, 
as well as robust animation under completely unseen poses. It can also generalize to novel identities, given only a short video as input. We demonstrate state-of-the-art performance for animating full-body talking humans, with fine-grained hand articulation and facial expressions. Project page: \url{https://aggelinacha.github.io/TalkinNeRF/}.
\keywords{Talking Humans \and Neural Radiance Fields \and Full-Body Animation}
\freefootnote{* This work was conducted while at Meta.}
\end{abstract}
\setlength{\belowcaptionskip}{-6pt}
\input{sec/1_intro}

\input{sec/2_related}
\input{sec/3_method}
\input{sec/4_experiments}
\input{sec/5_conclusion}
\clearpage
\bibliographystyle{splncs04}
\bibliography{References}

\input{suppl_arxiv}

\end{document}

%% file: sec/1_intro.tex
\section{Introduction}\label{sec:intro}
Synthesizing photorealistic 4D humans has been a long standing research problem in computer vision and graphics. It requires accurate capture of the spatio-temporal (4D) dynamics of the human body, including rigid and non-rigid deformations, such as facial expressions, with broad applications, ranging from AR/VR and video games to virtual communication and movies. Recent advances in dynamic neural radiance fields (NeRF)~\cite{mildenhall2020nerf} have enabled significant progress in rendering and animating human bodies~\cite{humannerf,peng2021neural} and faces~\cite{nerface,adnerf}. 
While many NeRF-based approaches require multiple views~\cite{peng2021neural,kwon2021neuralperf,park2021nerfies,park2021hypernerf,mu2023actorsnerf}, recent works learn dynamic NeRFs from monocular videos~\cite{nerface,humannerf,yu2023monohuman}. 
These works consider specific sub-problems, such as free-viewpoint rendering~\cite{humannerf}, rendering human subjects under novel poses~\cite{yu2023monohuman,mu2023actorsnerf}, and face reconstruction~\cite{nerface}. 
However, humans communicate with their full body, synchronizing their body pose, complex hand gestures, and facial expressions, in order to convey the intended message. Existing approaches do not consider this \emph{holistic} human motion, which is crucial for generating plausible digital talking humans.


In this work, we propose \MethodName, a novel method that learns a dynamic NeRF for \emph{full-body talking humans} from monocular videos. To the best of our knowledge, this is the first approach that introduces a \emph{unified} NeRF, combining body pose, hand articulation, as well as facial expressions, and learned from just monocular videos. 
In contrast to prior work, our problem setting presents the following challenges.
(a) Our training data consists of monocular \emph{frontal-only} videos of talking humans, whereas existing approaches leverage information from side and back views~\cite{li2021learn, peng2021neural, fang2021mirrored, dong2022totalselfscan, yu2023monohuman, mu2023actorsnerf, zheng2023avatarrex}. 
(b) We combine rigid and non-rigid human motion in a unified framework, where body parts correlate differently. 
(c) In talking humans, fine-grained details, such as finger articulation and facial expressions, play an important role in the final result, in order to convey the intended message in a conversation. Other works only consider the coarse limb articulation~\cite{mu2023actorsnerf,yu2023monohuman}. (d) We learn a \emph{multi-identity} representation, not identity-specific NeRFs as in prior work.

Given monocular videos of talking humans, our unified network represents their \emph{holistic} 4D human motion. For each video, we fit a parametric model and extract the parameters for the body pose, hand pose, and facial expression. These parameters condition corresponding modules for body, face, and hands, that are combined together to synthesize the final videos of \emph{full-body} talking humans. In order to capture complex finger articulation, we learn an additional deformation field for the hands. We also learn an identity code per subject, which enables us to train \MethodName on multiple identities simultaneously. Our multi-identity representation leverages information from the diverse motion of different subjects, enhancing the robustness under novel poses, while significantly reducing the overall training time. We also demonstrate animation guided by a speech-to-motion model, allowing for speech-to-video retargeting. Lastly, our method can generalize to an unseen identity, given a short video. 

In brief, our contributions are as follows:
\begin{itemize}
    \item We introduce \MethodName, a novel method that learns a unified dynamic NeRF for full-body talking humans from monocular videos, combining body pose, hand articulation, and facial expressions in a holistic manner.
    \item We propose a multi-identity representation, which enables us to simultaneously train on multiple identities, enhancing the robustness to novel poses, reducing the overall training time, and allowing for generalization to unseen identities given only short videos.
    \item \MethodName synthesizes high-quality videos of full-body talking humans, animating them under novel poses, with fine-grained hand articulation and facial expressions, outperforming the current state-of-the-art.
\end{itemize}

%% file: sec/2_related.tex
\section{Related Work}

\textbf{Neural Representations for Humans.}
Several methods for human body reconstruction and retargeting rely on parametric models~\cite{SMPL:2015,SMPL-X:2019,loper2023smpl,MANO:SIGGRAPHASIA:2017,AMASS:ICCV:2019,STAR:2020}, like SMPL~\cite{SMPL:2015}. These models are learned from large datasets of high-quality human scans. Then, they can be used as prior, conditioning neural representations of human avatars~\cite{humannerf,yu2023monohuman,mu2023actorsnerf,qian20233dgs}. Earlier methods for human body animation are based on image-to-image translation networks~\cite{pix2pix2017}. SMPLpix~\cite{smplpix} registers SMPL~\cite{SMPL:2015} to ground truth 3D scans. EverybodyDanceNow~\cite{everybodydancenow} extracts 2D skeletons~\cite{openpose} and learns an identity-specific image translation network. Subsequent works propose NeRF-based approaches~\cite{humannerf,nerface,dong2022totalselfscan,Feng2022scarf,shen2023xavatar,zheng2023avatarrex}, discussed in the following paragraphs. Most of them are identity-specific, can only represent one kind of motion (\eg, coarse rigid articulation or non-rigid facial motion), and require multiple views. Even recent works that propose techniques to reduce the training and inference times of radiance fields~\cite{zielonka2023insta,bakedavatar,qian20233dgs}, including 3DGS-based~\cite{qian2023gaussianavatars}, still have the same constraints. In contrast, our method combines the holistic rigid and non-rigid human motion in a unified framework, learned from in-the-wild monocular videos, and can generalize to multiple identities.

\noindent
\textbf{Neural Radiance Fields.}
Implicit neural representations have recently gained a lot of attention. NeRFs~\cite{mildenhall2020nerf,barron2021mipnerf,barron2022mipnerf360,lindell2022bacon} have been originally proposed for static scenes and have shown photorealistic novel view synthesis.
They represent a scene as a continuous 5D function, using a multilayer perceptron (MLP) that maps each 5D coordinate (3D spatial location and 2D viewing direction) to an RGB color and volume density. 
Recent works extend NeRFs to dynamic scenes~\cite{pumarola2021d,xian2021space,li2021neural,li2022neural,park2021nerfies,nerface,park2021hypernerf,humannerf}. They usually decouple 4D as 3D and time, and map (deform) the 3D points from an observation to a canonical space, in order to learn a time-invariant scene representation. Our work follows a similar concept, but we jointly represent rigid and non-rigid motion of talking humans.

\noindent
\textbf{Dynamic NeRFs for Humans.}
Deformations of the human face and body are particularly challenging to learn. Many NeRF-based approaches capture the 4D dynamics and appearance of humans from multi-view videos~\cite{peng2021neural,park2021nerfies,park2021hypernerf,xu2021h,2021narf,liu2021neuralactor,kania2022conerf,peng2023implicit,mpsnerf,jayasundara2023flexnerf}, while recent works use monocular videos~\cite{nerface,humannerf,yu2023monohuman,mu2023actorsnerf,jiang2022neuman,su2021nerf,wang2022neural,adnerf}. They align the observed poses of different video frames in a single canonical space, and learn a conditional 3D representation. By conditioning a NeRF on body pose~\cite{humannerf,yu2023monohuman}, hand pose~\cite{handnerf}, or facial expressions~\cite{nerface,rignerf,lipnerf} extracted from parametric models~\cite{SMPL:2015}, they enable meaningful control of the synthesized subject. They consider specific sub-problems, \ie, HumanNeRF~\cite{humannerf} shows remarkable results for free-viewpoint rendering, MonoHuman~\cite{yu2023monohuman} improves animation under novel poses, while NeRFace~\cite{nerface} focuses on face reconstruction and LipNeRF~\cite{lipnerf} on lip-syncing. However, all these works only animate the face, the hands, or the body separately. TotalSelfScan~\cite{dong2022totalselfscan} represents body, hand, and head pose, omitting the non-rigid motion of facial muscles, and requires multiple videos per subject. NeRFBlendshape~\cite{Gao2022nerfblendshape} and AD-NeRF~\cite{adnerf} represent only the head and face motion. X-Avatar~\cite{shen2023xavatar} requires 3D scans or RGB-D data in order to learn the geometry. AvatarReX~\cite{zheng2023avatarrex} requires multi-view videos - they use 22 cameras (16 for body and 6 for face). SCARF~\cite{Feng2022scarf} requires a detailed geometry (upsampled version of SMPL-X). In contrast, our method learns a unified network that represents the holistic 4D body motion of talking humans, given only frontal monocular videos.


\noindent
\textbf{Multi-Identity NeRFs for Humans.}
All the aforementioned approaches are identity-specific, requiring expensive optimization per identity. Only a limited number of prior work has aimed to train a generic NeRF for humans~\cite{raj2021pixel,hong2022headnerf,zhuang2022mofanerf,kwon2021neuralperf,mu2023actorsnerf,mpsnerf}. However, they all require multiple views for training. Multi-view approaches usually learn blending mechanisms, extract features from nearby views, and infer a novel view in feed-forward manner~\cite{wang2021ibrnet,chen2021mvsnerf,zheng2023gpsgaussian,kwon2021neuralperf}. Constrained by the input views, they cannot animate humans under novel poses. In contrast, in a monocular setting, we cannot leverage nearby views as input.
Another line of work~\cite{hu2023sherf,choi2022mononhr} learns a model from multi-view images, with the goal of rendering a new subject given only a single image as input. In contrast, we propose a simple architecture that is capable of learning multiple identities simultaneously, from just monocular videos, and can synthesize high-fidelity animations of them. We also demonstrate generalization to a new identity, given a short video.

%% file: sec/3_method.tex
\section{Method}

\begin{figure*}[t]
  \centering
   \includegraphics[width=\linewidth]{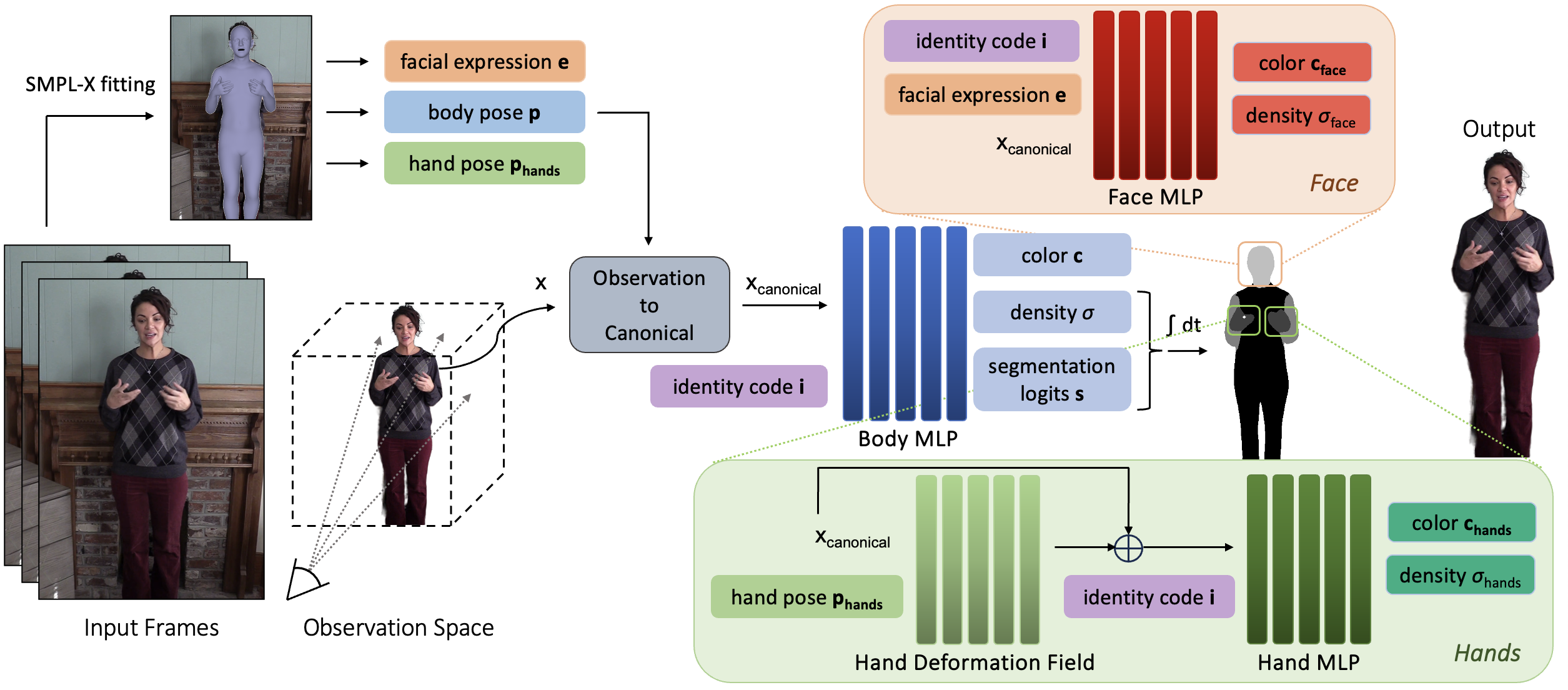}
   \caption{\textbf{Overview of \MethodName}. Given a monocular video of a subject, we learn a unified NeRF-based network that represents their holistic 4D motion. Corresponding modules for body, face, and hands are combined together, in order to synthesize the final \emph{full-body talking human}. By learning an identity code per video, our method can be trained on multiple identities simultaneously.}
   \label{fig:diagram}
\end{figure*}

We introduce \MethodName, a novel framework that learns a dynamic NeRF for full-body talking humans from monocular videos. 
An overview is illustrated in \cref{fig:diagram}. Given a monocular RGB video of a subject, we learn a unified network that represents their holistic 4D human motion.
For each video frame, we fit a parametric model and extract the parameters for the body pose, hand pose, and facial expression. These parameters condition corresponding modules for body, face, and hands, that are combined, to generate the final \emph{full-body} talking human. 
Additionally, we learn an identity code per video, which enables us to extend our unified representation to multiple identities. Leveraging information from multiple subjects, we enhance the robustness to unseen poses, while significantly reducing the total training time. \MethodName synthesizes high-quality videos of talking humans, with fine-grained hand articulation and facial expressions. It further generalizes to new identities, given only a short video of them.

\subsection{Conditional Representation}\label{sec:conditions}

\MethodName learns a conditional representation of the human body dynamics. First, for each video frame, we fit a parametric body model similar to SMPL-X~\cite{SMPL-X:2019},
and extract the estimated parameters for facial expression $\bm{\psi} \in \mathbb{R}^{10}$ and pose vectors for body $\bm{p_{\text{body}}} \in \mathbb{R}^{22 \times 3}$, jaw  $\bm{p_{\text{jaw}}} \in \mathbb{R}^{1 \times 3}$, and hands $\bm{p_{\text{hands}}} \in \mathbb{R}^{30 \times 3}$. 
The pose vectors are in axis-angles representation and describe the relative rotations of the joints. We also get the estimated camera intrinsics and extrinsics.

\noindent\textbf{Body Pose.} Based on the estimated parameters, we infer the relative joint locations, and convert the axis-angles representation to a $3 \times 3$ rotation matrix using the Rodrigues' rotation formula, following~\cite{humannerf}. In this way, we derive the corresponding rotation $\bm{R}_{i} \in \mathbb{R}^{{3 \times 3}}$ and translation $\bm{t}_{i} \in \mathbb{R}^{3}$ matrices for each body joint $i = 1, \dots, 22$. We define the body pose $\bm{p} = (\bm{R}, \bm{T})$, where $\bm{R} = \{\bm{R}_{i}\}$ and $\bm{T} = \{\bm{t}_{i}\}$ (see~\cite{humannerf} for the detailed derivation).

\noindent\textbf{Facial Expression.} We combine the expression coefficients $\bm{\psi}$ and jaw pose $\bm{p_{\text{jaw}}}$ to a vector $\bm{e} = [\bm{\psi}; \bm{p_{\text{jaw}}}] \in \mathbb{R}^{\text{13}}$. The jaw pose mostly captures the mouth openings and closures. Along with the expression coefficients $\bm{\psi}$ in the learned PCA space, they represent the facial expression $\bm{e}$ of the subject per frame.

\noindent\textbf{Hand Pose.} 
We directly concatenate the left and right hand poses (15 joints per hand) to a flattened vector $\bm{p_{\text{hands}}} \in \mathbb{R}^{\text{90}}$. 

\noindent
\textbf{Identity Code.} We additionally learn an identity code $\bm{i} \in \mathbb{R}^{32}$ per subject. This is a randomly initialized embedding that is learned during training. By enforcing it to be the same for all the frames of a subject, we ensure that it captures identity-specific information.

\subsection{Dynamic Neural Radiance Field}\label{sec:nerf}

We learn a unified representation $F_{\Theta}$ of the human motion, including body pose, hand articulation, and facial expressions. Following the dynamic NeRF literature~\cite{nerface,humannerf}, we consider our human subject in a 3D scene and learn an implicit representation $F_{\Theta}$ for each point in the scene. More specifically, given an identity $\bm{i}$ at a specific video frame, shown from a particular viewpoint and with a particular facial expression and body pose, we first march camera rays through the scene and sample 3D points on these rays. For a sampled 3D point $\bm{x}$, the estimated expression vector $\bm{e}$, body pose $\bm{p}$, and hand pose $\bm{p_{\text{hands}}}$, our learned $F_{\Theta}$ predicts the RGB color $\bm{c}$ and density $\sigma$ of the point:
\begin{equation}
    F_{\Theta}: (\bm{x}, \bm{e}, \bm{p},  \bm{p_{\text{hands}}}, \bm{i}) \longrightarrow (\bm{c}, \sigma)\;.
\end{equation}

\noindent\textbf{Canonical Space.} First, we map the points $\bm{x}$ from the observation space to the canonical space, similar to~\cite{humannerf}:
\begin{equation}\label{eq:canonical}
    \bm{x}_{\text{canonical}} = D_{\text{r}}(\bm{x}, \bm{p}) + D_{\text{nr}}(D_{\text{r}}(\bm{x}, \bm{p}), \bm{p})\;,
\end{equation}
where $D_{\text{r}}$ represents the rigid deformation of the joints and $D_{\text{nr}}$ accounts for any non-rigid deformations (\eg, due to clothing). $D_{\text{r}}$ corresponds to inverse linear blend skinning:
\begin{equation}\label{eq:rigid}
    D_{\text{r}}(\bm{x}, \bm{p}) = \sum_{i=1}^{K}w_c^i(\bm{x})(\bm{R}_i \bm{x} + \bm{t}_i)\;,
\end{equation}
where $K = 22$ is the number of body joints and $w_c^i$ is the blend weight for the $i$-th joint that is learned with a ConvNet during training~\cite{humannerf}. $D_{\text{nr}}$ corresponds to a trainable MLP, which adds a small offset to the skeleton-driven motion field, conditioned on the body pose.

\noindent
\textbf{Body MLP.} Each point $\bm{x}_{\text{canonical}}$ is first passed to the ``Body MLP'' $F_{\text{body}}$ (see \cref{fig:diagram}) that predicts its color $\bm{c}$, density $\sigma$, as well as segmentation logits $\bm{s}$ over $N$ classes:
\begin{equation}
    F_{\text{body}}: (\bm{x}_{\text{canonical}}, \bm{i}) \longrightarrow (\bm{c}, \sigma, \bm{s})\;,
\end{equation}
where $N = 5$ that include body, arms, hands, head, and background. We add the segmentation prediction as an additional output of the Body MLP for two main reasons: (a) we encourage the network to distinguish between the important body parts and pay more attention to the arm and hand gestures that are crucial while humans are conversing, and (b) we integrate additional modules for the face and hands, directly connected with the main Body MLP, avoiding any misalignment that can be caused by separate modules used in other works~\cite{adnerf}. When we predict that a point belongs to the hand or head regions, we pass it to the Hand or Face MLP correspondingly, described below.

\noindent
\textbf{Face MLP.} If $\bm{x}_{\text{canonical}}$ is categorized as head, we predict its color $\bm{c}$ and density $\sigma$ using the ``Face MLP''  $F_{\text{face}}$, conditioned on the corresponding facial expression $\bm{e}$:
\begin{equation}
    F_{\text{face}}: (\bm{x}_{\text{canonical}}, \bm{e}, \bm{i}) \longrightarrow (\bm{c}_{\text{face}}, \sigma_{\text{face}})\;,
\end{equation}
and set $\bm{c} = \bm{c}_{\text{face}}$ and $\sigma = \sigma_{\text{face}}$. In this way, our network learns non-rigid deformations caused by various facial expressions, and can synthesize expressive 3D humans.

\noindent
\textbf{Hand MLP.} If $\bm{x}_{\text{canonical}}$ is categorized as hands, we pass it to a hand deformation network $D_{\text{hands}}$, conditioned on the hand pose $\bm{p_{\text{hands}}}$. $D_{\text{hands}}$ adds an offset to the point location, in order to capture any deformations caused by the flexible movements of the finger joints:
\begin{equation}
    \bm{x}_{\text{canonical}} = \bm{x}_{\text{canonical}} + D_{\text{hands}}(\bm{x}_{\text{canonical}}, \bm{p_{\text{hands}}})\;.
\end{equation}
The output is passed to the ``Hand MLP''  $F_{\text{hands}}$ that predicts its color $\bm{c}$ and density $\sigma$:
\begin{equation}
    F_{\text{hands}}: (\bm{x}_{\text{canonical}}, \bm{i}) \longrightarrow (\bm{c}_{\text{hands}}, \sigma_{\text{hands}})\;.
\end{equation}
A key detail here 
is that unlike the face region, where we replace the color and density predicted by $F_{\text{body}}$ with the corresponding predictions of $F_{\text{face}}$, in the case of hands we use the predicted pixel color (accumulated colors and densities) given by $F_{\text{body}}$ as the color of the last point of the ray. In this way, we assume that the prediction of $F_{\text{body}}$ corresponds to the background of the hands, as they move around and can be on top of the subject's clothes or outside the human body. This ensures high-quality rendering, especially under novel poses (see ablation study in \cref{sec:exp_ablation}).

\noindent \textbf{Volume Rendering.}
Given the predicted color $\bm{c}$ and density $\sigma$ for every point on each ray, we produce the final video frame applying volume rendering~\cite{mildenhall2020nerf}. For each camera ray $\bm{r}(t) = \bm{o} + t\bm{v}$ with camera center $\bm{o}$ and viewing direction $\bm{v}$, the color $\hat{C}$ of the corresponding pixel can be computed by accumulating the predicted colors and densities of the sampled points along the ray:
\begin{equation}
    \hat{C}(\bm{r};\Theta) = \int_{t_n}^{t_f} \sigma (\bm{r}(t)) \bm{c} (\bm{r}(t)) T(t) dt\;,
\end{equation}
where $t_n$ and $t_f$ are the near and far bounds correspondingly, and $T(t) = \exp \left( - \int_{t_n}^{t} \sigma (\bm{r}(s)) ds \right)$ is the accumulated transmittance along the ray from $t_n$ to $t$. Similarly, we compute the segmentation logits $\hat{S}$ per image pixel:
\begin{equation}
    \hat{S}(\bm{r};\Theta) = \int_{t_n}^{t_f} \sigma (\bm{r}(t)) \bm{s} (\bm{r}(t)) T(t) dt\;.
\end{equation}

\noindent
\textbf{Optimization.} During training, we minimize the following objective function:
\begin{equation}
    \mathcal{L} = \mathcal{L}_{\text{LPIPS}} + \lambda_C \mathcal{L}_{C} + \lambda_S \mathcal{L}_{S}\;,
\end{equation}
\noindent where $\mathcal{L}_{C} = \sum_{\bm{r}} \left\| \hat{C}(\bm{r};\Theta) - C(\bm{r}) \right\|_{2}^{2}$ is the photometric loss that measures the pixel-wise difference between the ground truth color $C(\bm{r})$ and the predicted color $\hat{C}(\bm{r};\Theta)$ for all the rays $\bm{r}$, $\mathcal{L}_{\text{LPIPS}}$ is the perceptual loss LPIPS~\cite{zhang2018perceptual} using VGG as backbone, and  $\mathcal{L}_{S}$ is the categorical cross-entropy loss between the ground truth segmentation $S(\bm{r})$ and predicted $\hat{S}(\bm{r};\Theta)$. We set $\lambda_C = 0.2$ and $\lambda_S = 0.05$.

\noindent
\textbf{Implementation Details.}
We segment the human body from the background using an automatic human video matting method~\cite{rvm}. To get the segmentation classes, we use DensePose~\cite{guler2018densepose} and keep the corresponding labels for hands, arms, head, body (rest of body parts), and background. We use Adam optimizer~\cite{kingma2014adam} with a learning rate of $5\times10^{-4}$ with exponential decay, and train our network for 400k iterations on 4 GPUs (see suppl. for more details).


\subsection{Multi-Identity Optimization}\label{sec:multi_opt}

\MethodName can be trained using a single monocular video of a human subject or monocular videos of multiple subjects. 
By learning an identity code $\bm{i}$ per video, we extend our method to multiple identities, that can be trained simultaneously.
The identity code captures identity-specific information, \ie, the appearance of each identity, such as clothing, and conditions $F_{\text{body}}$, $F_{\text{face}}$, and $F_{\text{hands}}$. The network is fed with the diverse body poses, hand articulation, and facial expressions of all the training identities. By leveraging information from multiple subjects, 
\MethodName synthesizes high-quality videos of each of them under novel poses. Our multi-identity optimization reduces the overall training time by \(85\%\) - it only takes 600k iterations for 10 identities (compared to 400k iterations per identity). 

\noindent\textbf{Novel Identity.} \MethodName can also be adapted to a \textit{novel} identity, that is not part of the initial training set. Given a short video of a new subject, we fine-tune our network with a smaller learning rate ($10^{-5}$) for a few iterations (around 50k) to learn their identity code. We can then synthesize high-quality videos of them under novel poses.






%% file: sec/4_experiments.tex
\section{Experiments}

\noindent\textbf{Dataset.}
We collected 10 full-body talking human videos of different identities from the publicly available Casual Conversations dataset~\cite{hazirbas2021towards}. These videos are monocular and frontal-only. Each subject is standing and talking with their personal hand gestures and facial expressions. Only 2 subjects are walking, slightly showing their side to the camera, while talking. The videos are \mytilde1 minute long at 29.97 fps and $1080 \times 1080$ resolution. 
Compared to prior work~\cite{humannerf,yu2023monohuman}, our videos are more challenging for the following reasons: (a) they only have frontal views, whereas previous methods use videos with side and back views as well, (b) the variation in limb articulation is more limited, following a long-tailed distribution (our subjects are mostly standing and talking, with mainly arm and hand motion, in contrast to free-movement videos~\cite{li2021learn, peng2021neural,fang2021mirrored}), and (c) we include facial expression and hand articulation, 
compared to only body pose 
considered in prior work.

\noindent\textbf{Baselines.} We evaluate our method against state-of-the-art NeRF-based works, HumanNeRF~\cite{humannerf} and MonoHuman~\cite{yu2023monohuman}. We assess their performance when rendering a subject under novel poses and facial expressions.

\noindent\textbf{Evaluation Metrics.} We measure the visual quality of the generated videos, using peak signal-to-noise ratio (PSNR), structural similarity index (SSIM)~\cite{wang2004image}, and learned perceptual image patch similarity (LPIPS)~\cite{zhang2018perceptual}. Following related works, we report $\text{LPIPS}^{*} = \text{LPIPS} \times 10^3$.
Furthermore, we use the LSE-D (lip sync error - distance) and LSE-C (lip sync error - confidence) metrics~\cite{wav2lip,chung2016out}, to assess the lip synchronization, \ie, if the generated expressions are meaningful given the corresponding speech signal.



\begin{figure}[tb]
  \centering
  \begin{subfigure}{0.475\linewidth}
    \includegraphics[width=\linewidth]{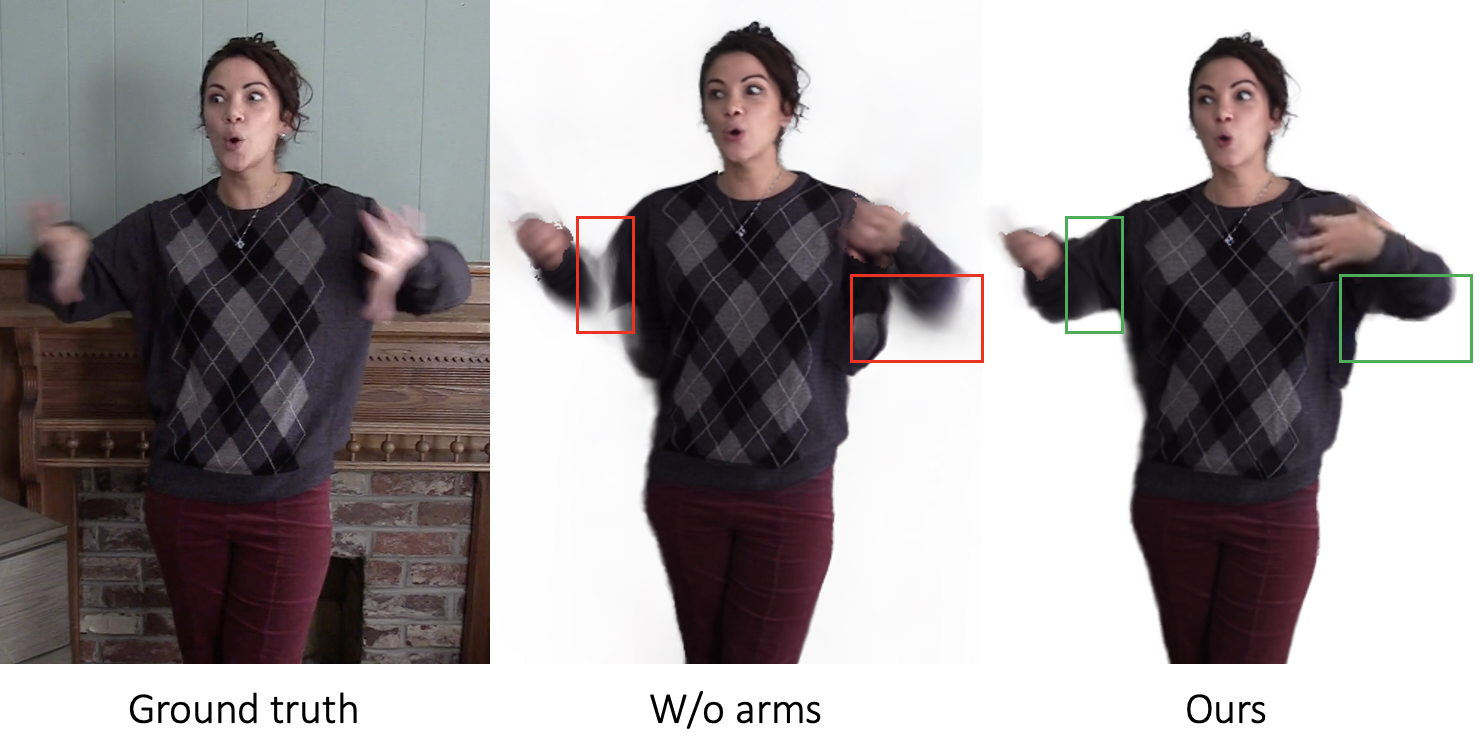}
   \caption{Ablation study on the segmentation classes}
   \label{fig:arms}
  \end{subfigure}
  \hfill
  \begin{subfigure}{0.475\linewidth}
    \includegraphics[width=\linewidth]{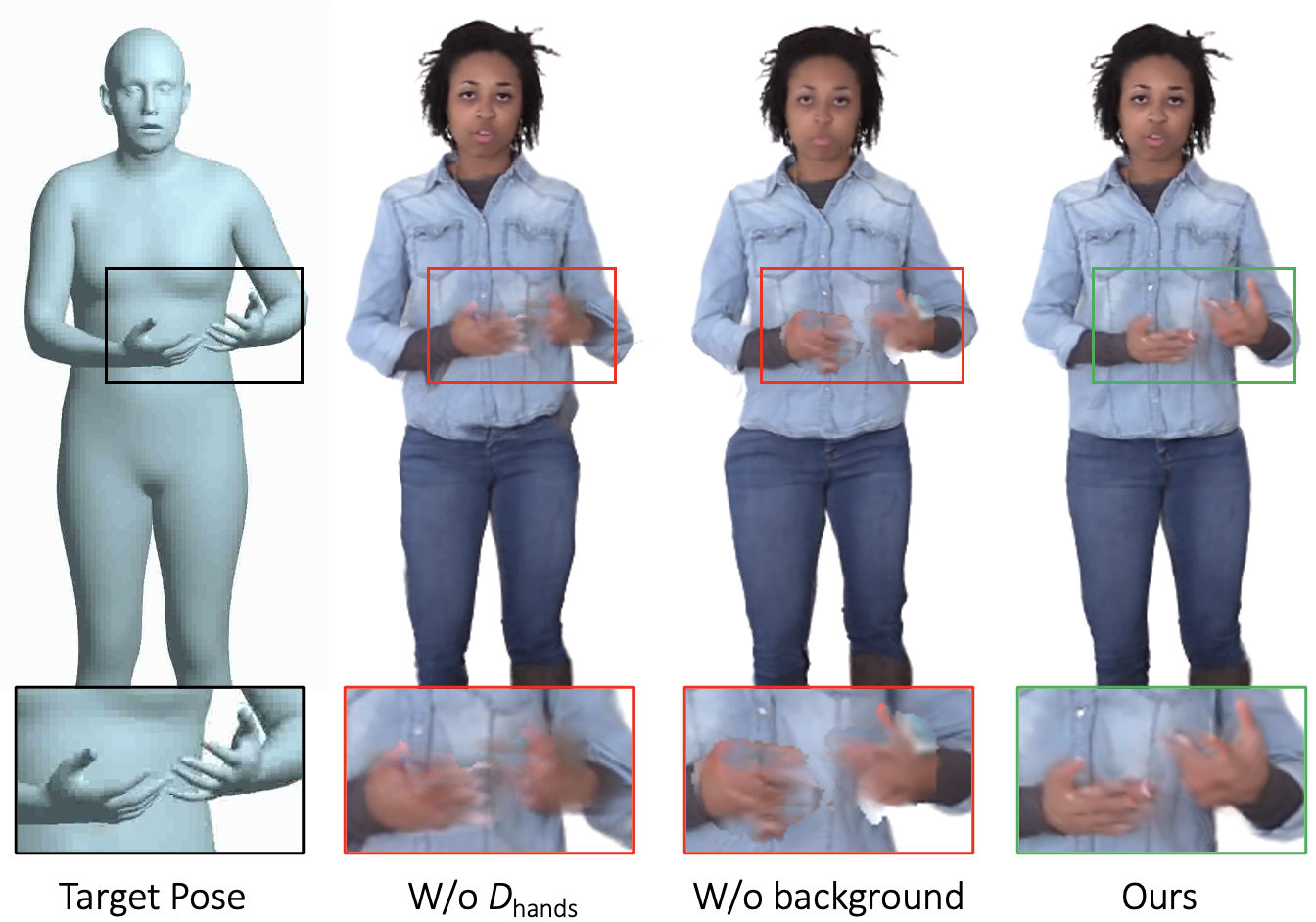}
   \caption{Ablation study on the hand representation}
   \label{fig:hands}
  \end{subfigure}
  \caption{\textbf{(a) Ablation study on the segmentation classes.} Our method predicts 5 classes: head, hands, arms, body, background. If the arms are considered as body (4 classes - ``W/o arms''), we observe disconnected arms in the reconstruction results. \textbf{(b) Ablation study on the hand representation when rendering novel (unseen) poses.} Without learning a hand deformation field $D_{\text{hands}}$, without considering the output of Body MLP as background for the hands (``W/o background''), and Ours (With $D_{\text{hands}}$ and background).}
\end{figure}

\begin{table}[t]
  \centering
  \caption{\textbf{Ablation study on the hand representation when rendering novel (unseen) poses.} We evaluate the following variants: without learning a hand deformation field $D_{\text{hands}}$, without considering the output of Body MLP as background for the hands (``W/o background''), when we pass each hand separately through the Hand MLP, when we include the hand joints in the observation-to-canonical transformation, and our complete approach. The metrics are computed only in the hand region.}
  \begin{tabular}{@{}l|ccc@{}}
    \toprule
    \textbf{Method Variant} & \textbf{$\text{PSNR}_{\text{}}$ $\uparrow$} & \textbf{$\text{SSIM}_{\text{}}$ $\uparrow$} & \textbf{$\text{LPIPS}^{*}_{\text{}}$ $\downarrow$} \\
    \midrule
    W/o hand deformation field & 17.56 & 0.4460 & 146.25 \\
    W/o background for hands & 17.07 & 0.3360 & 154.12 \\
    Separate left and right hands & 18.09 & 0.6282 & 87.09 \\
    Hand joints in canonical space & 17.98 & 0.6123 & 89.97 \\
    \hline
    Complete Approach & \textbf{18.15} & \textbf{0.6418} & \textbf{80.07}\\
    \bottomrule
  \end{tabular}
  \label{tab:ablation}
\end{table}

\begin{figure*}[t]
  \centering
   \includegraphics[width=\linewidth]{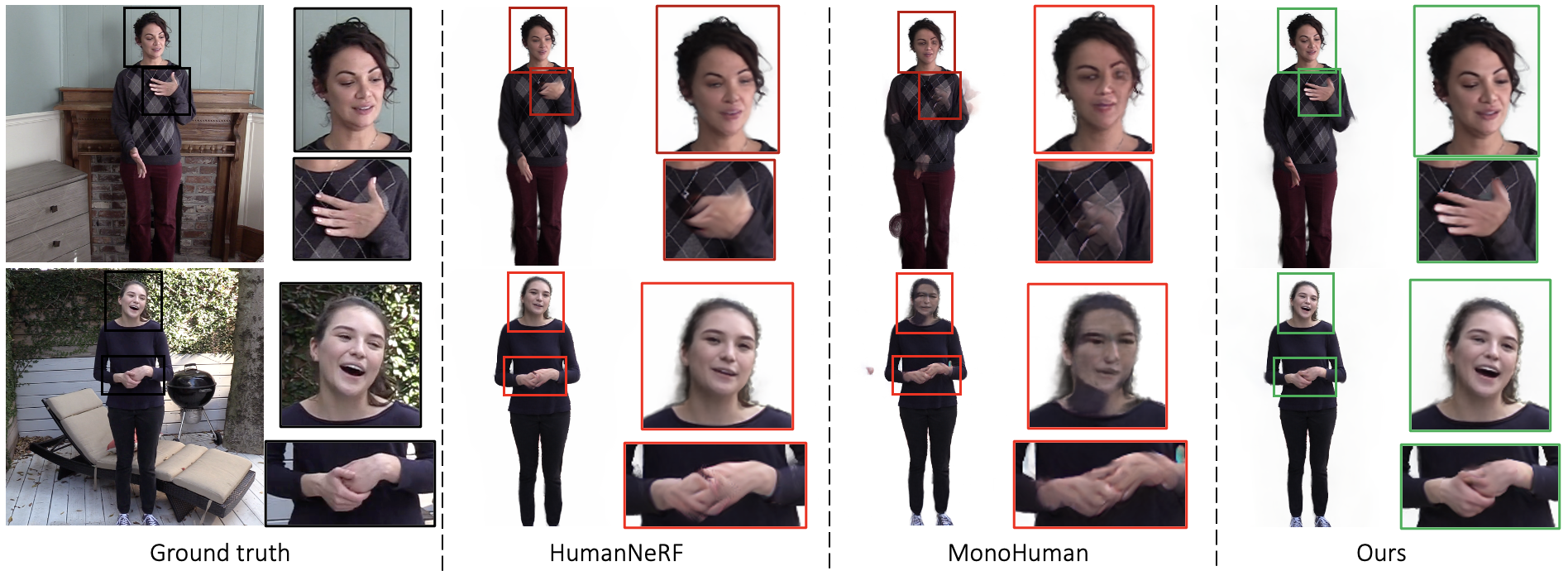}
   \caption{\textbf{Qualitative comparison for rendering novel poses from the same identity.} We compare with HumanNeRF~\cite{humannerf} and MonoHuman~\cite{yu2023monohuman}. Ground truth (not seen in training) is shown on the left. Our method generates facial expressions and hand articulation with a high fidelity.}
   \label{fig:reconstruction}
   \vspace{-0.2cm}
\end{figure*}

\begin{table}[t]
  \centering
  \caption{\textbf{Ablation study on the face representation.} We evaluate the variant that does not include the jaw joint. The metrics are computed only in the face region.}
  \label{tab:ablation_face}
  \begin{tabular}{@{}l|ccc@{}}
    \toprule
    \textbf{Method Variant} & \textbf{$\text{PSNR}_{\text{}}$ $\uparrow$} & \textbf{$\text{SSIM}_{\text{}}$ $\uparrow$} & \textbf{$\text{LPIPS}^{*}_{\text{}}$ $\downarrow$} \\
    \midrule
    W/o jaw for the face & 19.20 & 0.7532 & 33.50 \\
    \hline
    Complete Approach & \textbf{20.59} & \textbf{0.7677} & \textbf{31.48} \\
    \bottomrule
  \end{tabular}
  \vspace{-0.5cm}
\end{table}

\subsection{Ablation Study}\label{sec:exp_ablation}
We conduct an ablation study on the architecture of \MethodName. First, we investigate how the segmentation prediction of the Body MLP affects our model. As described in \cref{sec:nerf}, the hand and head classes are necessary for the proposed pipeline, in order to pass the corresponding points to the hand or face-related modules. We observe that adding the arms as a separate class encourages the network to pay attention to the subject's gestures and avoids any disconnected arms from the main body, due to fast motion (see \cref{fig:arms}).

An important part of our method is the representation of the hands. \cref{fig:hands} and \cref{tab:ablation} demonstrate the results of different variants of our hand representation when rendering a specific subject under novel poses. 
By just using the Hand MLP, the network can overfit to the training poses and render them correctly. However, synthesizing novel poses is much more challenging. Without learning a hand deformation field $D_{\text{hands}}$, the network cannot generate a plausible deformation of the points caused by the various finger movements (\cref{fig:hands} column 2, \cref{tab:ablation} row 1). If we do not consider the output of the Body MLP as background for the hands, the predicted hand color might erroneously include nearby points, leading to artifacts (\cref{fig:hands} column 3, \cref{tab:ablation} row 2). 
We also investigate a variant where we pass each hand separately through the Hand MLP (\cref{tab:ablation} row 3). However, we observe a small increase in performance when we concatenate both hands together. We suspect that this happens because the hands are highly symmetrical and interact with each other.
Thus, their combined pose is a more informative input to the model. We also consider the case where we include the hand joints in the transformation from the observation to the canonical space, \ie,~ in \cref{eq:canonical,eq:rigid} (\cref{tab:ablation} row 4). Although mapping the hand joints to the canonical space might be useful when animating just the hands~\cite{handnerf}, in our case, that we animate body and hands jointly, this degrades the visual quality under novel poses. We suspect that this happens because of an increase in the number of transformations and in the input size of $D_{\text{nr}}$, discouraging the network to pay attention to the main body parts.
Lastly, for the face representation, as shown in \cref{tab:ablation_face}, the jaw joint provides additional information for the mouth motion.

\begin{figure*}[t]
  \centering
   \includegraphics[width=\linewidth]{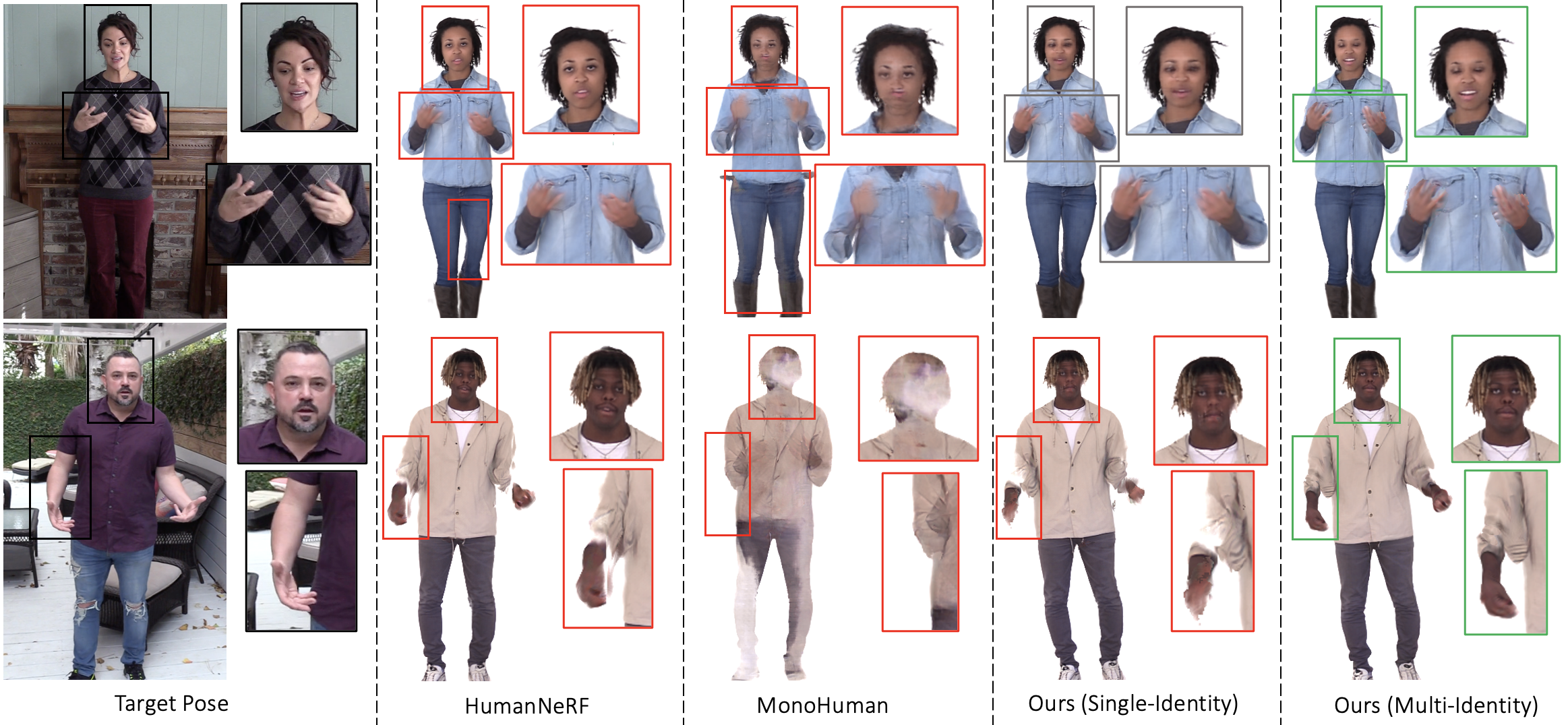}
   \caption{\textbf{Qualitative comparison for rendering novel poses from a different identity.} From left to right: target pose, results of HumanNeRF~\cite{humannerf}, MonoHuman~\cite{yu2023monohuman}, our single-identity model, and our multi-identity model. Our multi-identity \MethodName robustly renders each identity under unseen poses and expressions.}
   \label{fig:novelposes_expose}
\end{figure*}



\subsection{Evaluation: Novel Poses}\label{sec:exp_novelposes}
 We identify 3 main evaluation settings, with increasing difficulty: rendering a subject under (a) poses from the same identity, (b) poses from a different identity, and (c) speech-driven generated poses.

\noindent
\textbf{Poses from the same identity.} We train our model on the 10 full-body talking videos and evaluate it on a test set of held-out frames per identity. \cref{fig:reconstruction} demonstrates the corresponding qualitative results. \MethodName synthesizes the facial expression and fine-grained finger articulation with a high fidelity. 
HumanNeRF~\cite{humannerf} can only learn an average facial expression that is the same for all the frames, which is a major limitation for talking humans. MonoHuman~\cite{yu2023monohuman} fails to learn a good representation of the human subjects, as it needs a selection of front and back frames that do not exist in our data. This is an unrealistic assumption for several application scenarios, including talking humans that naturally talk in front of the camera without turning their back. 
\Cref{tab:rec} shows the corresponding quantitative results on the held-out frames. Our method can be trained on each identity separately, or on all the training videos simultaneously. By learning from multiple identities, \MethodName enhances the overall performance under novel poses and is trained significantly faster (\mytilde85\% faster than training a single-identity model per subject). 
However, in this evaluation setting, we noticed that sometimes the hands may be more faithfully produced by the single-identity model (slight decrease in LPIPS for the hands). 
We suspect that this happens because each subject's hands can vary in size, skin color, and articulation, and the single-identity model can better memorize identity-specific details when rendering poses from the same identity. This is not true though when rendering completely novel poses from other identities, as shown next.

\begin{table}[t]
  \centering
    \caption{\textbf{Quantitative evaluation for rendering novel poses from the same identity.} We compare HumanNeRF~\cite{humannerf}, MonoHuman~\cite{yu2023monohuman}, our single-identity, and multi-identity models, computing PSNR, SSIM, LPIPS on the full image, and LPIPS only in the hand region ($\text{LPIPS}^{*} = \text{LPIPS} \times 10^3$). Our multi-identity \MethodName outperforms the other methods.}

  \begin{tabular}{@{}l|cccc@{}}
    \toprule
    \textbf{Method} & \textbf{PSNR $\uparrow$} & \textbf{SSIM $\uparrow$} & \textbf{$\text{LPIPS}^{*}$ $\downarrow$} & \textbf{$\text{LPIPS}^{*}_{\text{hands}}$ $\downarrow$}\\
    \midrule
    HumanNeRF & 25.36 & 0.9108 & 38.62 & 132.54\\
    MonoHuman & 24.94 & 0.9010 & 47.40 & 116.91\\
    \hline
    Ours (Single-Id) & 25.78 & 0.9132 & 33.21 & \textbf{80.07}\\
    Ours (Multi-Id) & \textbf{27.25} & \textbf{0.9341} & \textbf{32.06} & 82.27\\
    \bottomrule
  \end{tabular}
  \label{tab:rec}
\end{table}

\begin{table}[t]
  \centering
    \caption{\textbf{Quantitative evaluation for rendering novel facial expressions.} We compare HumanNeRF~\cite{humannerf}, MonoHuman~\cite{yu2023monohuman}, our single-identity, and multi-identity models, computing LSE-D and LSE-C metrics. Our multi-identity \MethodName outperforms the other methods.}
  \begin{tabular}{@{}l|cccc@{}}
    \toprule
    \textbf{Method} & \textbf{LSE-D $\downarrow$} & \textbf{LSE-C $\uparrow$}\\
    \midrule
    HumanNeRF & 11.42 & 0.200\\
    MonoHuman & 11.01 & 0.157\\
    \hline
    Ours (Single-Id) & 10.23 & 1.552 \\
    Ours (Multi-Id) & \textbf{9.54} & \textbf{2.712}\\
    \bottomrule
  \end{tabular}
  \label{tab:lse}
\end{table}

\begin{figure}[t]
  \centering
   \includegraphics[width=0.7\linewidth]{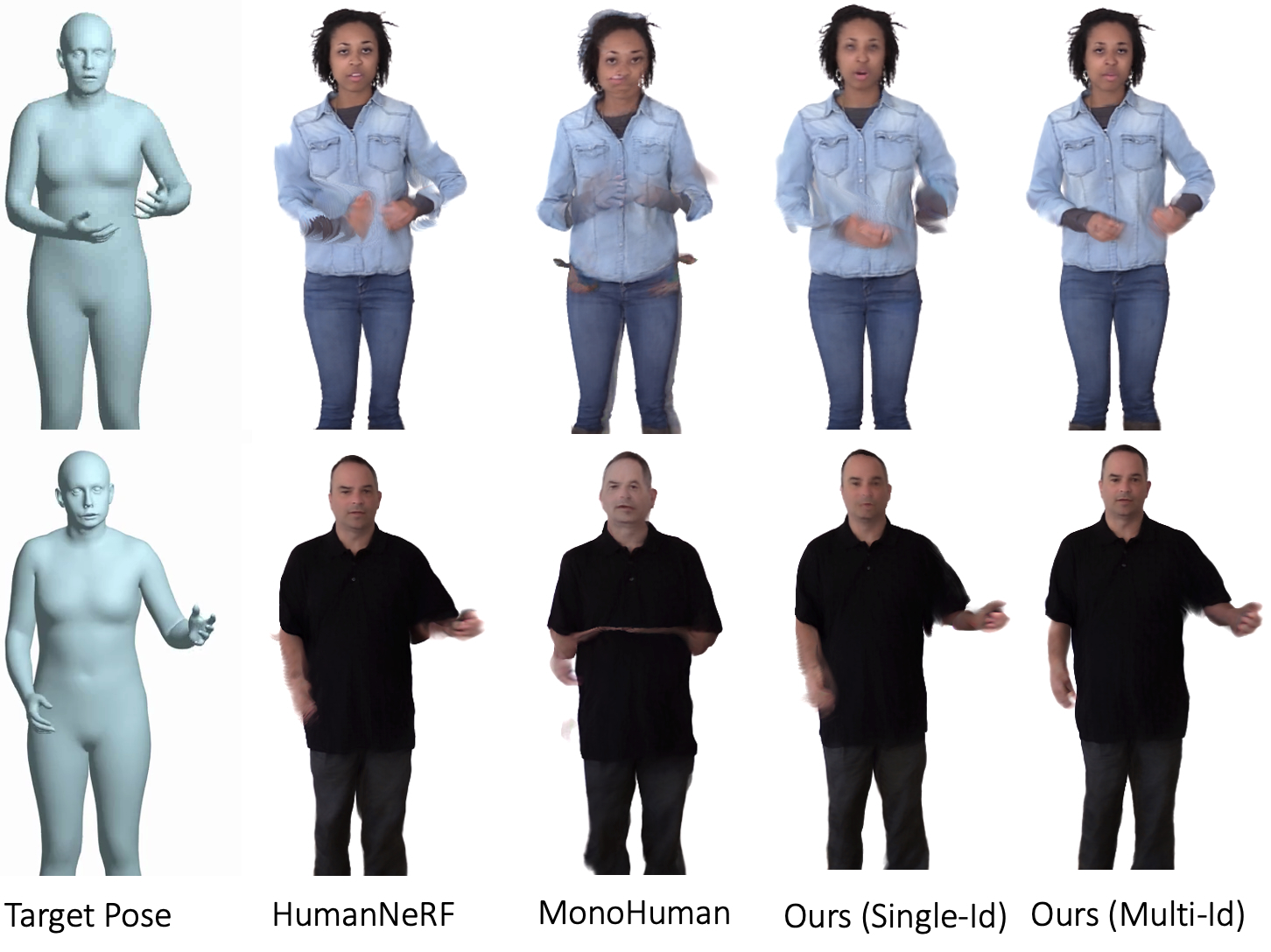}
   \caption{\textbf{Qualitative comparison for rendering unseen out-of-distribution poses}. From left to right: target pose, results of HumanNeRF~\cite{humannerf}, MonoHuman~\cite{yu2023monohuman}, our single-identity, and our multi-identity model. Our multi-identity \MethodName robustly renders each identity under completely unseen poses and expressions.}
   \label{fig:novelposes_talkshow}
\end{figure}

\noindent
\textbf{Poses from other identities.} Animating a subject under novel poses from different identities is more challenging. Each subject has their own personal talking style and hand articulation. Thus, novel poses from a different subject might be far from the training distribution. Our multi-identity representation significantly outperforms the other methods in this setting, as demonstrated in the qualitative results in \cref{fig:novelposes_expose}. 
Notice in the first row, how it can photo-realistically synthesize an unseen facial expression with closed eye lids and a smile, following the target expression. In addition, it robustly produces the unseen hand and arm pose, as well as facial expression in the second row.
Since we do not have ground truth in this case, we use the LSE-D and LSE-C metrics~\cite{wav2lip,chung2016out} to evaluate the generated facial expressions, given the corresponding speech signal (see \cref{tab:lse}). 

\noindent
\textbf{Speech-driven generated poses.} 
Another interesting setting is rendering poses generated from audio prompts. In this case, we use a speech-to-motion model~\cite{ng2024audio} that synthesizes sequences of 3D human body poses, including hand gestures and facial expressions, given a speech recording or a text-to-speech output. 
These generated poses are more challenging, since they use different data for training and they follow a different fitting algorithm. \cref{fig:novelposes_talkshow} demonstrates the corresponding qualitative results. HumanNeRF, MonoHuman, and the single-identity model can lead to artifacts (\eg, disconnected arms and hands), while our multi-identity \MethodName robustly produces high-quality animations that faithfully follow the target poses. 

\noindent



\subsection{Evaluation: Novel Identity}\label{sec:exp_novelid}

\begin{figure}[t]
  \centering
  \begin{subfigure}{0.45\linewidth}
    \includegraphics[width=\linewidth]{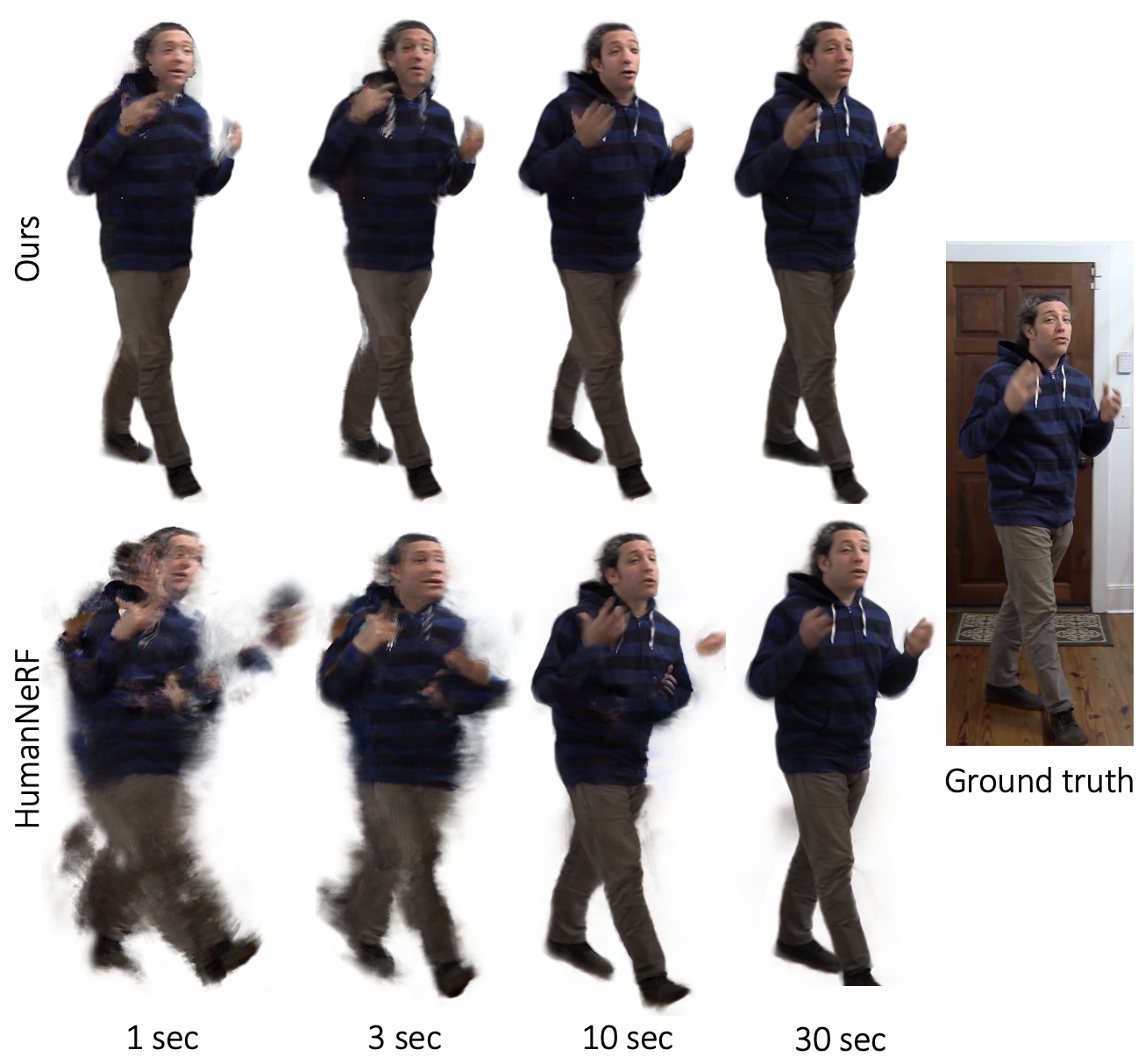}
   \caption{}
   \label{fig:few}
  \end{subfigure}
  \hfill
  \begin{subfigure}{0.45\linewidth}
    \includegraphics[width=\linewidth]{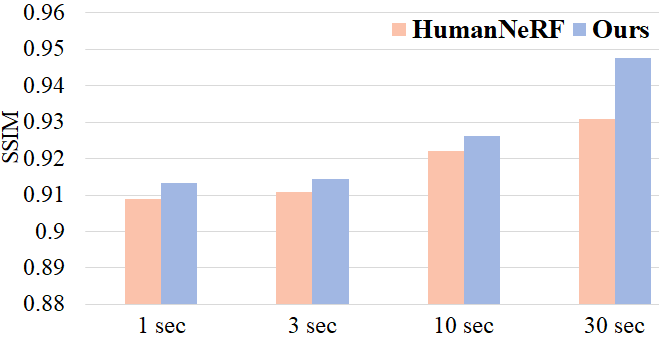}
   \caption{}
   \label{fig:few_bar}
  \end{subfigure}
  \caption{\textbf{Learning a novel identity.} \textbf{(a) Qualitative comparison.} The novel identity is learned from a short video (1, 3, 10 or 30 sec.), \ie, small number of consecutive frames. Compared to HumanNeRF~\cite{humannerf}, our method robustly renders the new identity under completely unseen poses, given only a very short video. \textbf{(b) Quantitative comparison.} The novel identity is learned from a short video (1, 3, 10 or 30 sec.).} 
\end{figure}

We also explore the scenario when we would like to animate a \textit{novel} identity, which originally is not part of our training subjects. Given only a short video, \ie, a small number of consecutive frames, we evaluate if our multi-identity model can learn to robustly animate this new identity. Note that in this case, only a very small variation of body poses, hand gestures, and facial expressions is seen for that particular subject. We initially train our multi-identity \MethodName using the 9 full-body talking videos, and then adapt it to the \(10^{th}\) identity, as described in~\cref{sec:multi_opt}, using very small portions of their videos (1, 3, 10, or 30 seconds). \cref{fig:few} demonstrates the corresponding results when rendering a novel identity under novel poses, \emph{completely unseen in training}. Our method robustly animates a novel identity, even given a very short video, in contrast to HumanNeRF. Corresponding quantitative results are illustrated in \cref{fig:few_bar}. 

\begin{figure}[t]
  \centering
    \includegraphics[width=0.85\linewidth]{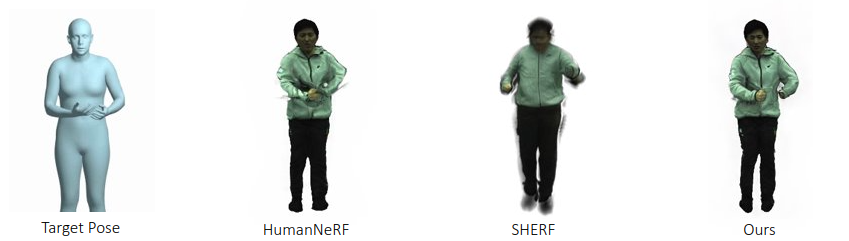}
  \caption{\textbf{Learning a novel identity.} From left to right: target pose, results of HumanNeRF~\cite{humannerf}, SHERF~\cite{hu2023sherf}, and our multi-identity model. Our method robustly renders the new identity under completely unseen poses.} 
  \label{fig:sherf}
\end{figure}

\cref{fig:sherf} shows additional qualitative comparison with SHERF~\cite{hu2023sherf}, which produces 3D humans from a single input frame. Given a short video of a new identity,
we fine-tune our multi-identity model, in order to learn the corresponding identity code. We train HumanNeRF~\cite{humannerf} and fine-tune SHERF~\cite{hu2023sherf} using the same short video. \MethodName significantly outperforms the other methods, robustly animating the completely new identity under completely unseen poses. 



\subsection{Limitations}
Similarly with related works, the fitting algorithm that estimates the input body poses, hand articulation and facial expressions can be noisy. These are used as ground truth for training and thus, any error is propagated to our method and the final results.
It might introduce temporal inconsistency. In addition, it might not be able to capture the exact finger positions in fast moving frames and subtle facial movements. In the future, we plan to collect higher resolution data for the face and hands to further enhance the visual quality.

%% file: sec/5_conclusion.tex
\section{Conclusion}
We present \MethodName, a novel approach that represents the holistic 4D motion of talking humans from monocular videos. 
To the best of our knowledge, this is the first approach to introduce a dynamic NeRF that combines body pose, hand articulation, as well as facial expression, learned from monocular frontal-only videos. We introduce a multi-identity representation that enables us to simultaneously train on multiple subjects. In this way, not only we reduce the overall training time, but also we enhance our method's robustness under completely unseen poses and expressions.
\MethodName obtains state-of-the-art performance in animating full-body talking humans. We provide several applications, ranging from adapting to a completely new identity, given only a short video, to generating a retargeted motion from a speech-to-motion model.




%% file: suppl_arxiv.tex
\clearpage
\setcounter{section}{0}
\renewcommand\thesection{\Alph{section}}

\section*{Supplementary}

\noindent
The supplementary document is organized as follows: 
\begin{enumerate}
    \item Additional Results
    \item Implementation Details
\end{enumerate}
We strongly encourage the readers to watch our supplementary video on our project page: \url{https://aggelinacha.github.io/TalkinNeRF/}.

\section{Additional Results}

\begin{figure*}[t]
  \centering
   \includegraphics[width=0.9\linewidth]{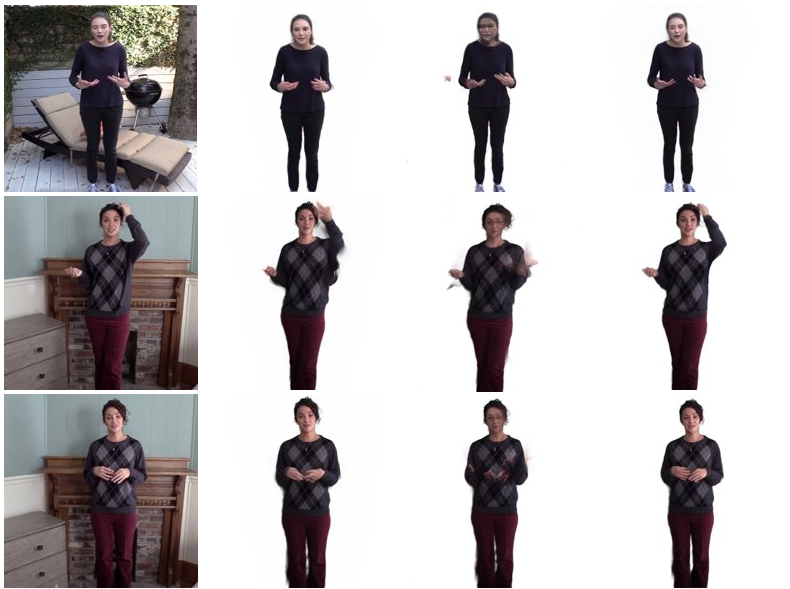}
   \includegraphics[width=0.9\linewidth]{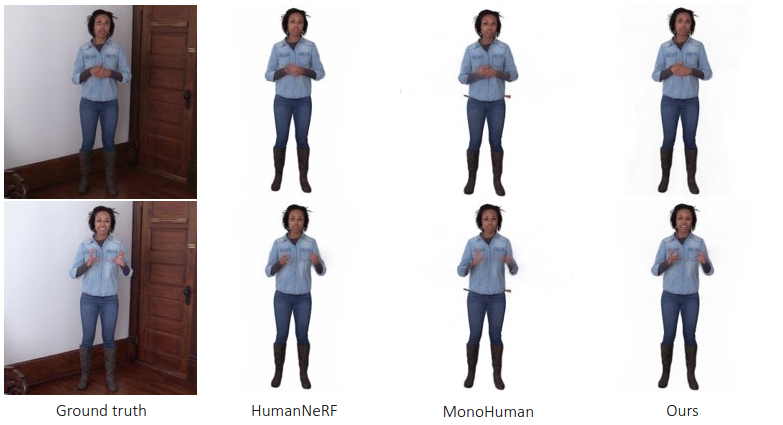}
   \caption{\textbf{Qualitative comparison for rendering novel poses from the same identity.} We compare with HumanNeRF~\cite{humannerf} and MonoHuman~\cite{yu2023monohuman}. Ground truth (not seen in training) is shown on the left. Our method generates facial expressions and hand articulation with a high fidelity.}
   \label{fig:rec_suppl}
\end{figure*}

\cref{fig:rec_suppl} shows additional qualitative results when rendering novel poses from the same identity, not seen during training. Please notice the facial expressions and the hands of each subject. \MethodName synthesizes them with a high fidelity. As also mentioned in Sec.~4.2, HumanNeRF~\cite{humannerf} learns only an average expression per subject, while MonoHuman~\cite{yu2023monohuman} produces artifacts, since our data include only frontal videos.

\begin{figure*}[t]
  \centering
   \includegraphics[width=\linewidth]{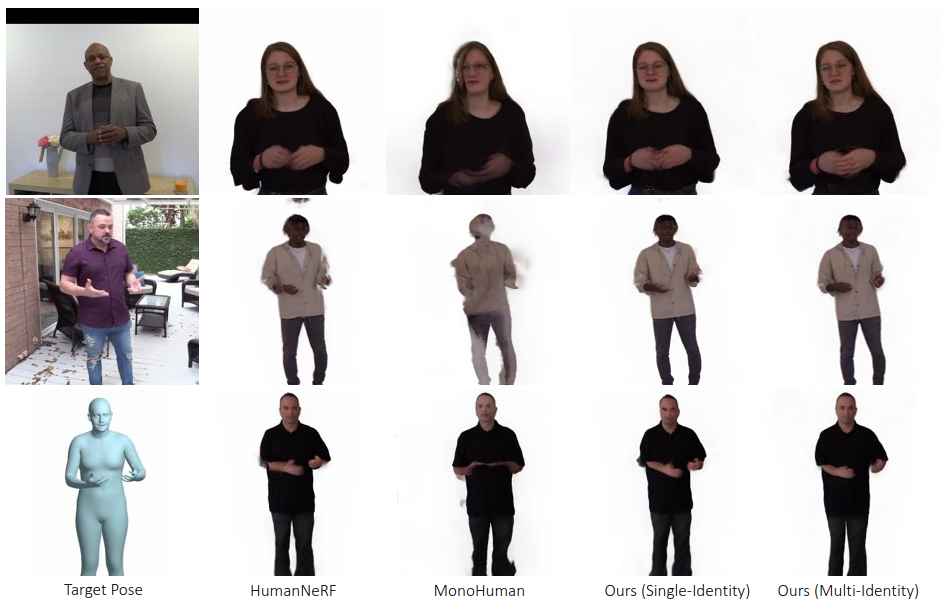}
   \caption{\textbf{Qualitative comparison for rendering novel (unseen) poses.} From left to right: target pose, results of HumanNeRF~\cite{humannerf}, MonoHuman~\cite{yu2023monohuman}, our single-identity model, and our multi-identity model. Our multi-identity \MethodName robustly renders each identity under unseen poses and expressions.}
   \label{fig:novel_suppl}
\end{figure*}

\cref{fig:novel_suppl} shows additional qualitative results when rendering novel poses from different identities (first 2 rows), or speech-driven poses generated by a speech-to-motion model~\cite{yi2023generating} (last row). We compare HumanNeRF~\cite{humannerf}, MonoHuman~\cite{yu2023monohuman}, our single-identity, and our multi-identity model. In contrast to the other methods, our multi-identity \MethodName robustly renders each identity under completely unseen body poses, hand articulation, and facial expressions.

\begin{figure*}[t]
  \centering
   \includegraphics[width=\linewidth]{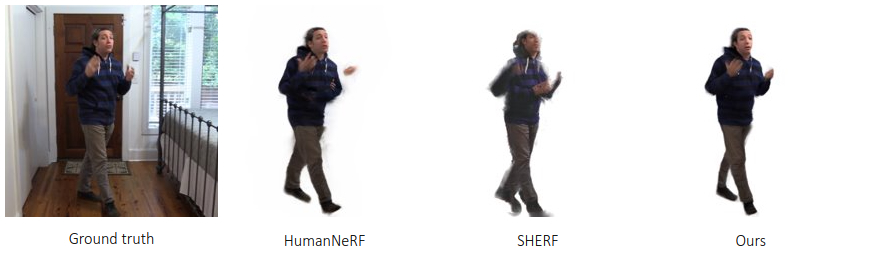}
   \caption{\textbf{Qualitative comparison for learning a novel identity.} From left to right: ground truth, results of HumanNeRF~\cite{humannerf}, SHERF~\cite{hu2023sherf}, and our multi-identity model. Our method robustly renders the new identity under completely unseen poses, given only a very short video of 10 seconds.}
   \label{fig:sherf_suppl}
\end{figure*}

\cref{fig:sherf_suppl} shows additional qualitative results when rendering a novel identity, which originally is not part of our training subjects. As mentioned in Sec.~4.3, given a short video of the new identity, we fine-tune our multi-identity model, in order to learn the corresponding identity code. Since we only use a few consecutive frames, our model only sees a very small variation of body poses, hand gestures, and facial expressions of the new identity. Similarly, we train HumanNeRF~\cite{humannerf} and fine-tune SHERF~\cite{hu2023sherf} using the same short video. \MethodName significantly outperforms the other methods, robustly animating the new identity under completely unseen poses.

\begin{figure}[t]
  \centering
    \includegraphics[width=0.7\linewidth]{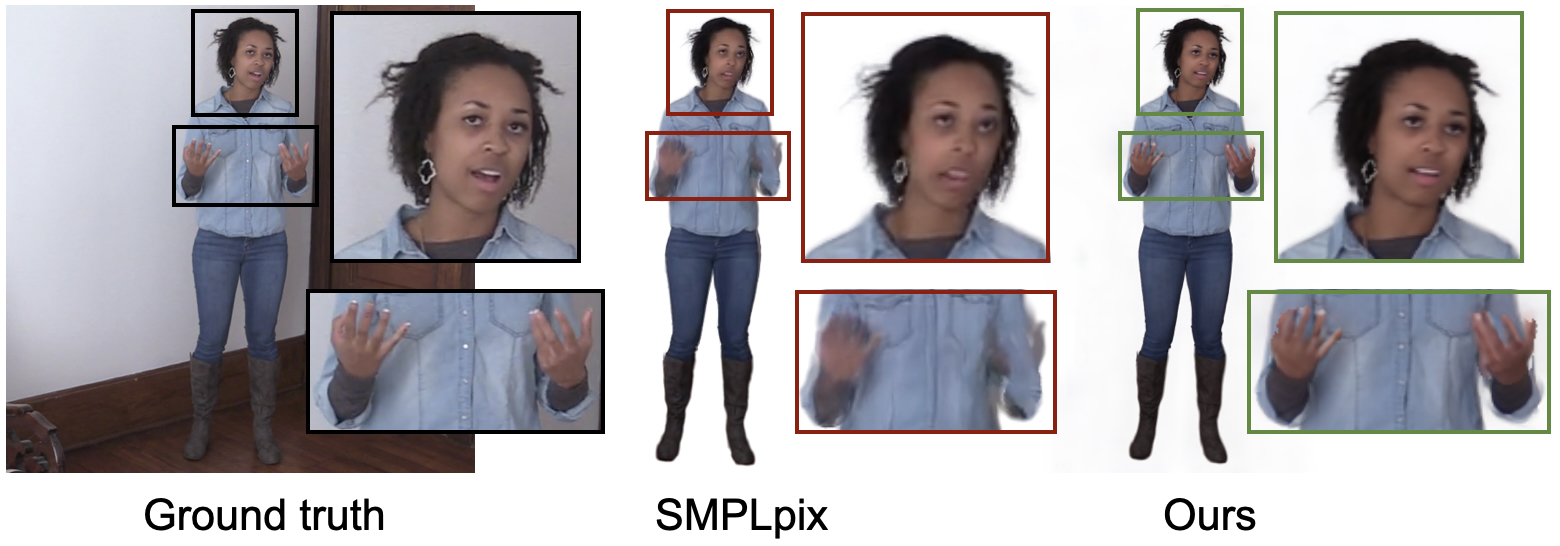}
   \caption{Qualitative comparison with SMPLpix~\cite{smplpix}. Our method captures fine-grained facial expression and hand articulation.}
    \label{fig:smplpix}
\end{figure}

\begin{figure}[t]
  \centering
  \includegraphics[width=0.8\linewidth]{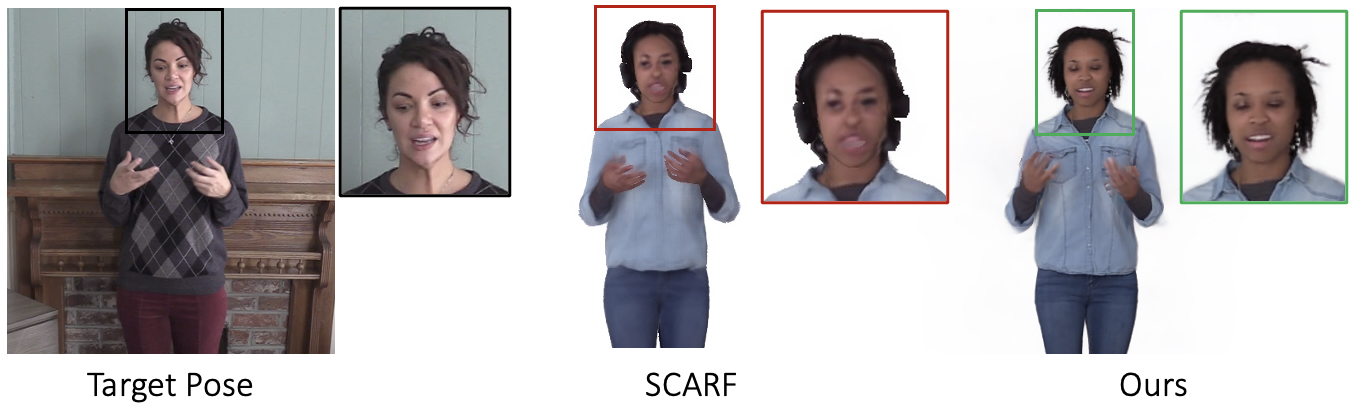}
   \caption{Qualitative comparison with SCARF~\cite{Feng2022scarf}. Our method synthesizes high visual quality, whereas SCARF leads to blurry faces.}
   \label{fig:scarf}
\end{figure}

\cref{fig:smplpix} compares our method with SMPLpix~\cite{smplpix}, which learns an image-to-image translation network for rendering. SMPLpix fails to synthesize photo-realistic facial expressions and hands. In \cref{fig:scarf}, we further compare with SCARF \cite{Feng2022scarf} that leads to blurry faces. Our method produces high-quality videos of talking humans.


\noindent
\textbf{Video Results.} We strongly encourage the readers to watch our supplementary video that shows animations of full-body talking humans, and corresponding comparisons with the state-of-the-art.

\section{Implementation Details}

In this section, we include additional implementation details. 

\noindent
\textbf{Segmentation.} As mentioned in Sec.~3.2, we segment the human body from the background using an automatic human video matting method~\cite{rvm}. To get the segmentation ground truth classes, we use DensePose~\cite{guler2018densepose} and keep the corresponding labels for hands, arms, head, body (rest of body parts), and background. 

\noindent
\textbf{Architecture.} For the body pose, we closely follow the architecture of HumanNeRF~\cite{humannerf}, in order to map the points from the observation to the canonical space and train the Body MLP. For $F_{\text{body}}$, $F_{\text{face}}$, and $F_{\text{hands}}$, we use MLPs of 8 linear layers with 256 hidden units each, ReLU activations, and sinusoidal positional encodings with 10 frequency bands for the input points. For $D_{\text{nr}}$ and $D_{\text{hands}}$, we use MLPs of 6 linear layers with 128 hidden units each, ReLU activations, and positional encodings with 6 frequency bands with a truncated Hann window applied~\cite{humannerf}. To avoid overfitting to the seen poses for each subject, $D_{\text{nr}}$ is included in the training after 100k iterations.

\noindent
\textbf{Sampling.} We march 32768 rays and sample 128 points per ray during training. We use stratified sampling~\cite{mildenhall2020nerf} and sample points mostly inside the 3D bounding box of the human subject, with a percentage of 80\%. Since arm and hand gestures are important in talking humans, and only cover a small area in the video frames, we additionally set certain sampling percentages for arms and hands. Based on our ground truth segmentation, we sample 20\% of the points inside the bounding box from the hands during the first 20k iterations, 20\% from the hands and 60\% from the arms during the next 20k iterations, 40\% from the hands and 40\% from the arms during the next 20k, 80\% from the hands and 20\% from the arms during the next 20k, and 20\% from the hands for all the following iterations. We empirically found that this sampling strategy encourages the network to learn the structure and motion of arms and hands at appropriate iterations for our data.

\noindent
\textbf{Training.}
Our implementation is based on PyTorch~\cite{paszke2019pytorch}. We train our single-identity network for 400k iterations and our multi-identity network for 600k iterations for 10 subjects, using 4 GPUs.
We use Adam optimizer~\cite{kingma2014adam} with an initial learning rate of $5\times10^{-4}$ and exponential decay, with a rate of 0.1 and 500k steps. The rest of the Adam hyper-parameters are set at their default values 
($\beta_1 = 0.9, \beta_2 = 0.999, \epsilon = 10^{-8}$). To compute the LPIPS loss, we apply patch-based ray sampling, similarly with~\cite{humannerf}. We sample 6 patches with a size of $32 \times 32$ on each image.

\noindent
\textbf{Rendering.}
Since the target poses are usually noisy, estimated in a frame-by-frame manner~\cite{ExPose:2020}, we apply temporal smoothing for our final video synthesis during test time. We use a Gaussian filter with a window size of 5 frames and unit standard deviation, applied to the target body poses, hand poses, facial expressions and camera parameters, along the temporal axis. This ensures temporally smooth results.

\noindent
\textbf{Evaluation.}
We compute the visual quality metrics (PSNR, SSIM, LPIPS) for the full video frames, only the face, and only the hands, as indicated in our quantitative results.
To compute them for the face region, we crop each frame around the face, using the face detector from 3DDFA~\cite{guo2020towards,3ddfa_cleardusk}. Similarly for the hand region, to evaluate the hand visual quality, we crop a bounding box around each hand, based on the segmentation by DensePose~\cite{guler2018densepose}.